\documentclass[letterpaper]{article}

\usepackage{iclr2027_conference,times}
\iclrfinalcopy

\usepackage{graphicx}
\usepackage{caption}
\usepackage{booktabs}
\usepackage{multirow}
\usepackage{array}

\usepackage{amsmath}
\usepackage{amsfonts}

\usepackage{algorithm}
\usepackage{algorithmic}

\usepackage[hyphens]{url}
\usepackage{hyperref}
\usepackage{xcolor}
\usepackage{colortbl}
\usepackage{tabularx}
\definecolor{RefineFlyRow}{RGB}{232,242,250}
\newcommand{\best}[1]{\textbf{#1}}
\newcommand{\second}[1]{\underline{#1}}

\newcommand{\meanres}[1]{{\normalsize #1}}
\newcommand{\stdres}[1]{{\small\ensuremath{\pm\,#1}}}

\title{RefineFly: Failure-Aware Post-Training for
Aerial Vision-Language Navigation}

\author{
    Boxiong Wang\textsuperscript{1,4},
    Hui Kang\textsuperscript{1},
    Geng Sun\textsuperscript{1}%
    \thanks{Corresponding authors: Geng Sun and Jiahui Li.}\\
    Jiahui Li\textsuperscript{1}\footnotemark[1],
    Chao Yu\textsuperscript{2,4},
    Daxin Tian\textsuperscript{3,4}\\[0.5em]
    \textnormal{
        \textsuperscript{1}Jilin University
        \quad
        \textsuperscript{2}Tsinghua University
    }\\
    \textnormal{
        \textsuperscript{3}Beihang University
        \quad
        \textsuperscript{4}Zhongguancun Academy
    }\\
    \textnormal{
        \texttt{wangbx0320@163.com,
        \{sungeng,lijiahui\}@jlu.edu.cn}
    }
}

\begin{document}

\maketitle
\lhead{Preprint}

\begin{abstract}
Unmanned aerial vehicle vision-language navigation (UAV-VLN) requires agents to translate visual observations and language instructions into reliable flight actions in complex environments. Although recent end-to-end UAV vision-language-action (UAV-VLA) policies reduce reliance on separately designed perception, planning, and control modules, their behavior-cloning objectives provide limited supervision for errors arising during closed-loop execution. Reinforcement learning (RL) offers a promising solution, while limited interaction budgets, long-tailed scene distributions, and policy drift complicate sustained learning. To this end, we propose RefineFly, a failure-aware RL post-training framework for end-to-end UAV-VLA policies. Built on token-level proximal policy optimization (PPO), RefineFly maintains dynamic failure memory within a two-stage scene curriculum to sustain learning from unresolved tasks as training shifts from empirical to balanced scene sampling. Throughout this process, stage-specific reference regularization constrains policy deviation. Experiments on the TravelUAV benchmark demonstrate that RefineFly outperforms all comparison methods across all three splits, improving success rate over AerialVLA by 3.12 to 8.37 percentage points with a total rollout budget of about 30\% of the training-set size. Moreover, ablations reveal that the benefits of sustained failure learning vary across evaluation splits, while the two-stage scene curriculum improves overall performance, highlighting the importance of training distributions when learning from failures.
\end{abstract}

\section{Introduction}

\par Vision-language navigation (VLN) requires embodied agents to associate natural-language instructions with visual observations and execute trajectories toward semantic goals \citep{Anderson2018,Krantz2020}. Unmanned aerial vehicle (UAV) VLN extends this problem to large-scale three-dimensional (3D) environments, where an aerial agent searches for a language-described object while continuously adjusting forward motion, altitude, yaw, and termination decisions \citep{aerialvln,Fan2023,traveluav}. Recent vision-language-action (VLA) models provide a promising end-to-end interface from onboard observations and instructions to executable controls, reducing reliance on separately engineered perception, planning, and control modules \citep{Zitkovich2023,openvla,aerialvla}.

\par However, behavior-cloned UAV-VLA policies receive limited supervision for errors arising during their own closed-loop execution. Existing UAV-VLN systems improve long-horizon navigation through waypoint prediction, hierarchical planners, memory mechanisms, or external perception modules \citep{traveluav,Zhang2025,Ding2026HETT,Ning2026LookasideVLN,longfly}. Although end-to-end policies such as AerialVLA \citep{aerialvla} generate flight actions directly, their training relies on expert demonstrations that provide limited coverage of error-induced states \citep{Ross2011}. Consequently, rapidly changing viewpoints, occlusion, and long trajectories can cause local errors to compound into collisions, navigation drift, premature landing, or timeouts. These failures expose diverse forms of unsuccessful behavior and provide feedback for further policy improvement. Accordingly, online reinforcement learning (RL) post-training offers a way to refine the pretrained policy using rewards from its own navigation outcomes.

\begin{figure}[t]
\centering
\includegraphics[width=0.76\linewidth]{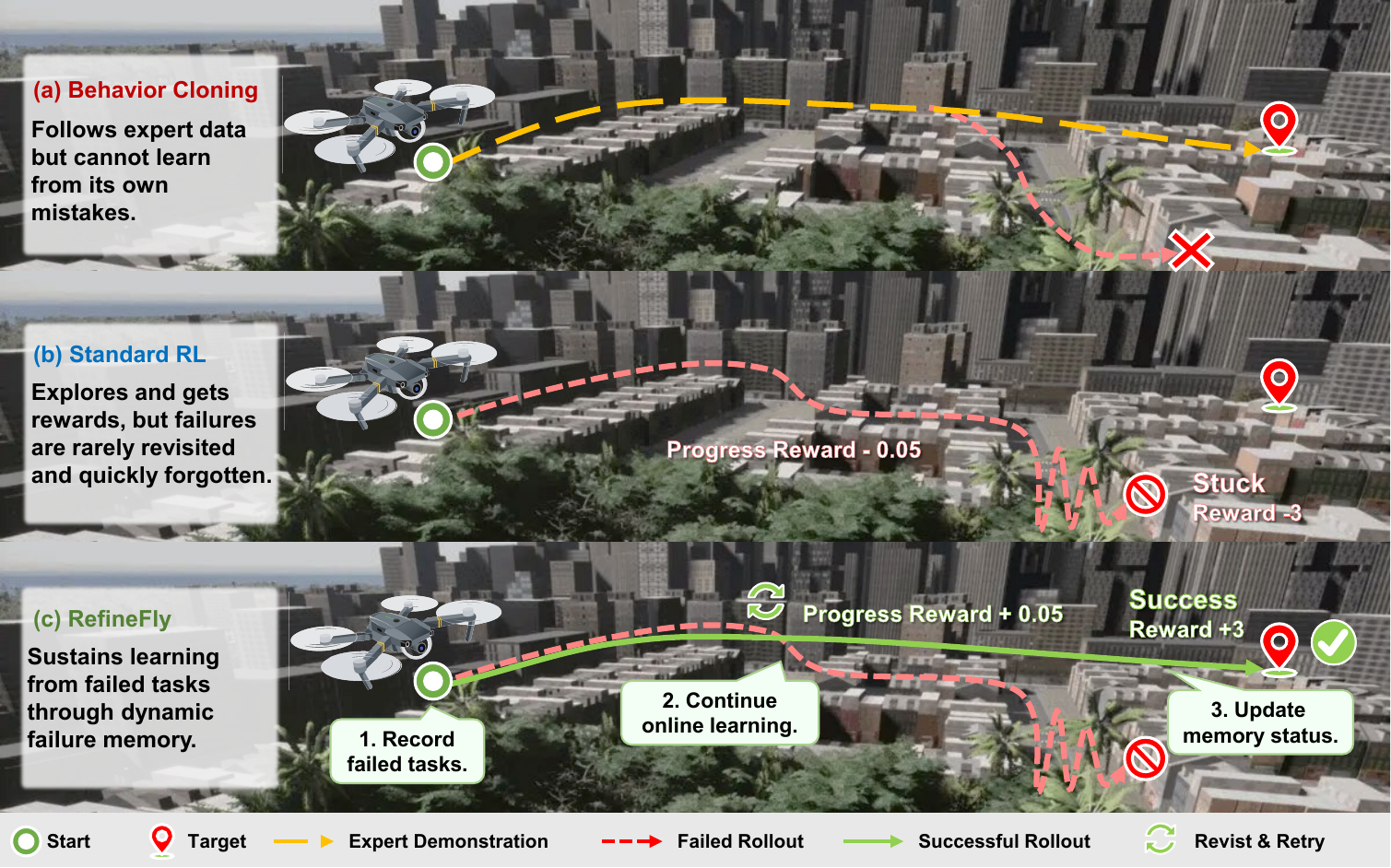}
\caption{\textbf{Comparison of training paradigms for UAV-VLN.} RefineFly sustains online learning from unresolved tasks through dynamic failure memory, with fresh rollouts informing both policy updates and the memory lifecycle.}
\label{fig: overview}
\end{figure}

\par Nevertheless, exploiting this feedback effectively requires careful organization of post-training. Specifically, long navigation rollouts involve repeated simulation and VLA inference, making extensive online interaction costly. Within a limited budget, ordinary task sampling may provide few opportunities to continue learning from unresolved failures. Moreover, uneven scene distributions further restrict these opportunities in environments with relatively few training tasks \citep{lin2025openvln,gao2025openfly}. Consequently, post-training needs to reconcile adaptation under the existing task distribution with broader training coverage of rare scenes. Meanwhile, changes in the selected failure cases and scene distribution can shift the focus of policy updates, potentially weakening previously acquired navigation capabilities. These challenges motivate sustained failure learning together with stage-wise distribution adaptation and constraints on policy deviation.

\par To address these challenges, we propose RefineFly, a failure-aware online post-training framework for end-to-end UAV-VLA policies, as illustrated in Fig.~\ref{fig: overview}. Built on token-level proximal policy optimization (PPO) \citep{Schulman2017}, RefineFly sustains learning from unresolved tasks through dynamic failure memory within a two-stage scene curriculum. As training shifts from empirical to balanced scene sampling, the framework expands learning opportunities for rare scenes while constraining policy deviation through reference regularization. This design supports continued learning from closed-loop failures and navigation generalization under a finite online interaction budget.

\par The main contributions of this paper are as follows.
\begin{itemize}
\item \textbf{RL Post-Training for End-to-End UAV-VLA.} We adapt RL post-training to end-to-end VLA policies for UAV-VLN and empirically study the influence of training distributions and sampling strategies in online post-training, together with the benefits and trade-offs of sustained learning from failure experiences.

\item \textbf{Failure-Aware Online Post-Training Framework.} We propose RefineFly, which integrates dynamic failure memory with lifecycle management into a two-stage scene curriculum built on token-level PPO. This design sustains learning from failures across stage-specific training distributions while using reference regularization to mitigate policy drift.

\item \textbf{Navigation Performance and Generalization.} Experiments demonstrate that RefineFly outperforms the compared baselines across all evaluation splits, with substantial gains on challenging long-horizon navigation tasks.
\end{itemize}

\section{Related Work}

\subsection{VLN for UAVs}
\par Early VLN methods established cross-modal instruction grounding and action prediction \citep{Anderson2018,Krantz2020}. AerialVLN \citep{aerialvln} and AVDN \citep{Fan2023} extended this setting to outdoor aerial navigation and dialog-conditioned target search, while UAV-ON \citep{xiao2025uav} and OpenFly \citep{gao2025openfly} broadened it toward open-world object goals and larger-scale aerial datasets. To execute long-horizon flights, TravelUAV \citep{traveluav} predicts continuous waypoints, whereas CityNavAgent \citep{zhang2025citynavagent}, SkyVLN \citep{li2025skyvln}, TypeFly \citep{Chen_2025}, and training-free VLM approaches \citep{hu2025see} combine language reasoning with memory, trajectory generation, model-based control, or program synthesis. Moreover, NavFoM \citep{Zhang2025} and LongFly \citep{longfly} further improve generalization through large-scale pretraining and spatiotemporal modeling. These methods substantially improve semantic planning, while most retain explicit interfaces between high-level reasoning and low-level flight execution, through which prediction and control errors can accumulate.

\subsection{VLA Policies for UAV-VLN} 
\par RT-2 \citep{Zitkovich2023} and OpenVLA \citep{openvla} show that robotic actions can be represented in language-compatible output spaces, allowing pretrained vision-language representations to support embodied control. This paradigm has been extended to fine-grained imitation, racing, and cognitive UAV control \citep{wang2026uav,serpiva2025racevla,lykov2025cognitivedrone}. Furthermore, AerialVLA \citep{aerialvla} maps onboard observations and linguistic prompts directly to continuous 3-degree-of-freedom (DoF) controls and an intrinsic landing decision, reducing reliance on oracle waypoints, external detectors, and separate landing modules. However, these UAV-VLA policies are trained mainly by supervised behavior cloning, which provides limited coverage of states induced by autonomous execution errors \citep{Ross2011}. Consequently, closed-loop failures motivate further policy improvement through online feedback.

\subsection{RL Post-Training for Aerial VLN}
\par Interactive RL post-training has been explored directly in language-conditioned UAV control. SuReAL \citep{Blukis2020UAV} combines supervised position prediction with RL-based continuous control and evaluates the resulting policy on a physical quadcopter. More recently, HTNav \citep{Fan2026HTNav} integrates staged imitation and RL with tiered decision making for urban aerial VLN. OpenVLN \citep{lin2025openvln} studies policy adaptation under data scarcity using value-based waypoint rewards and KL-regularized updates, while FlightGPT \citep{cai2025flightgpt} combines supervised fine-tuning with reward-based refinement for goal accuracy, reasoning quality, and output compliance. These studies establish the value of interaction feedback, reward design, and staged optimization for improving aerial navigation. Building on this direction, our work studies how sustained learning from failed tasks interacts with scene sampling during end-to-end UAV-VLA post-training. Accordingly, RefineFly maintains dynamic failure memory throughout a two-stage scene curriculum to support continued policy improvement under uneven scene coverage.

\section{Method}
\subsection{Overview}

\par
\begin{figure}[t]
\centering
\includegraphics[width=\textwidth]{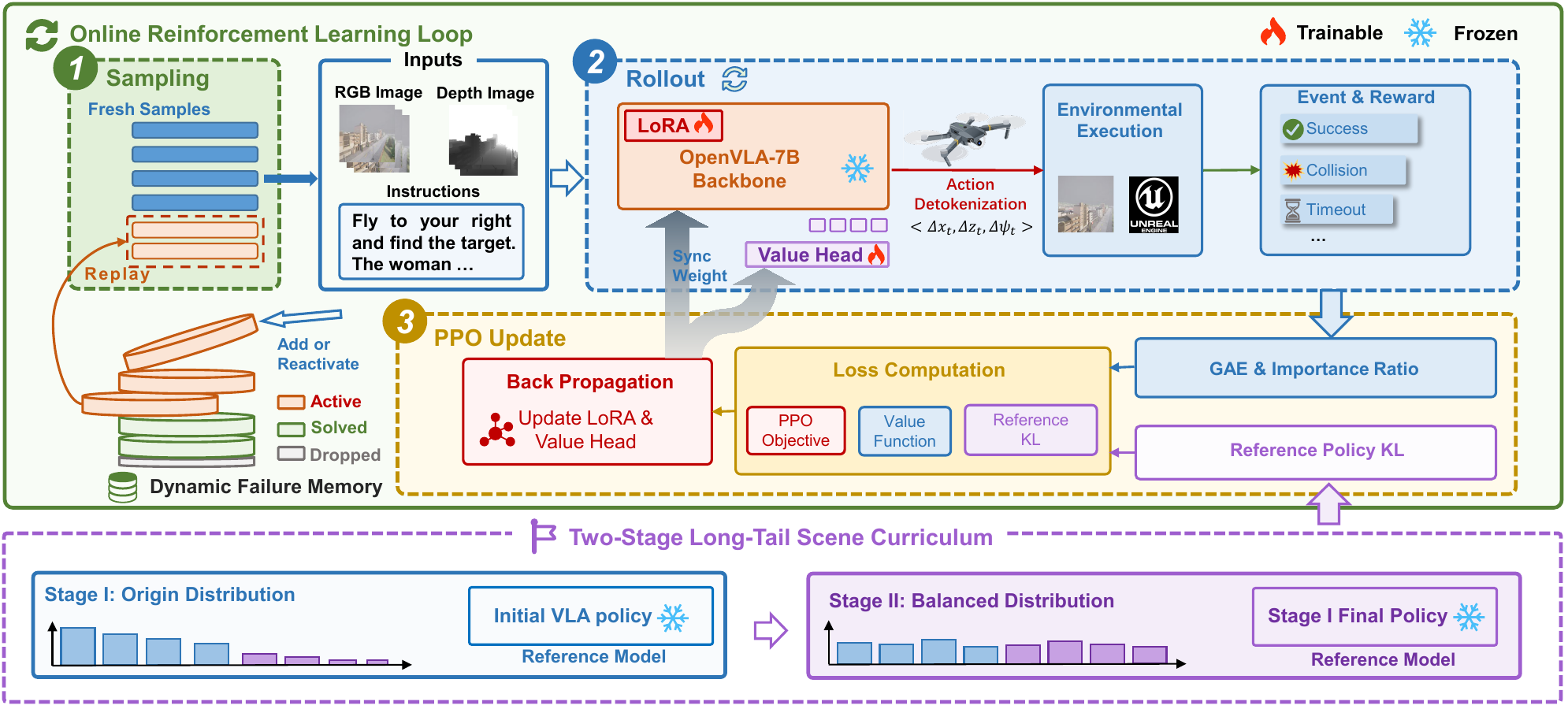} 
\caption{\textbf{Overview of RefineFly.}
RefineFly sustains learning from unresolved tasks through dynamic failure memory within a two-stage curriculum that rebalances the training distribution across scenes. Token-level PPO updates the policy through the original autoregressive textual action interface, with updates regularized against a reference policy frozen within each stage.}
\label{fig: architecture}
\end{figure}

 We propose RefineFly, a failure-aware online RL post-training framework for end-to-end VLA policies in UAV-VLN, as illustrated in Fig.~\ref{fig: architecture}. Using token-level PPO as the optimization backbone, RefineFly preserves the original autoregressive textual action interface. During online learning, the framework maintains a dynamic failure memory to sustain learning from unresolved tasks while retaining fresh task sampling from the underlying training distribution. Building on this process, a two-stage curriculum shifts training from the original long-tailed distribution toward a more balanced distribution, providing more training opportunities for tasks in rare scenes and enabling further learning from the associated failures. Throughout this process, policy updates are regularized against a reference policy frozen within each stage to mitigate policy drift as the failure memory and training distribution evolve.

\subsection{Policy Interface and Optimization Backbone}
\paragraph{Textual Action Interface.}
\par We consider end-to-end UAV-VLN control and adapt the UAV-VLA policy to RL fine-tuning through its existing autoregressive textual action interface. Specifically, at time step $t$, the instruction $x$ and visual observation $o_t$ form the context $h_t=(x,o_t)$, from which the policy generates an action-token sequence $\mathbf z_t=(z_{t,1},\ldots,z_{t,K_t})$ of length $K_t$ according to
\begin{equation}
\pi_\theta(\mathbf z_t\mid h_t)
=\prod_{k=1}^{K_t}\pi_\theta(z_{t,k}\mid h_t,z_{t,<k}).
\label{eq:textual_policy}
\end{equation}
Following AerialVLA~\citep{aerialvla}, three uniformly quantized control tokens represent forward progression, vertical movement, and yaw adjustment, followed by an optional \texttt{LAND} token. A deterministic decoder then maps valid sequences to continuous commands $\mathbf a_t=\langle\Delta x_t,\Delta z_t,\Delta\psi_t\rangle$.

\par Variable-length action-token sequences require PPO updates to account for both valid token positions and executable command formats.
Specifically, unmasked padding can contribute to the loss, and unnormalized token sums can overweight longer commands. Accordingly, we pad sequences to a maximum length $K$, apply a mask $m_{t,k}\in\{0,1\}$ to select valid action-token positions, and normalize the loss by $M_t=\max(1,\sum_{k=1}^{K}m_{t,k})$. Meanwhile, the same action grammar constrains generation and likelihood computation. Any remaining parsing failure produces a safe no-op and receives an invalid-action penalty.

\paragraph{Token-Level PPO Backbone.}
\par Sequence-level PPO applies a shared clipping decision to the joint probability ratio of an autoregressive action sequence. Consequently, a probability change at one token position can activate clipping and suppress policy-gradient contributions from other positions. Accordingly, we adopt token-level PPO as the optimization backbone, applying probability ratios and clipping separately at each valid action-token position. Here, all valid tokens within a command share the action-level advantage $\widehat A_t$. For the rollout policy $\pi_{\mathrm{old}}$, the token probability ratio is
\begin{equation}
\rho_{t,k}(\theta)=\frac{\pi_\theta(z_{t,k}\mid h_t,z_{t,<k})}{\pi_{\mathrm{old}}(z_{t,k}\mid h_t,z_{t,<k})}.
\end{equation}

\par To incorporate closed-loop navigation feedback into token-level updates, we compute the shared action-level advantage from rollout rewards and value estimates. During training, a value head $V_\phi(h_t)$ provides these estimates and is optimized using the standard clipped PPO value loss $\mathcal L_V$. With $\bar\rho_{t,k}(\theta)=\operatorname{clip}(\rho_{t,k}(\theta),1-\epsilon,1+\epsilon)$, the masked policy objective is
\begin{equation}
J_{\mathrm{PPO}}^{\mathrm{token}}(\theta)
=\mathbb E_t\!\left[
\frac{1}{M_t}\sum_{k=1}^{K}m_{t,k}
\min\!\left(
\rho_{t,k}(\theta)\widehat A_t,
\bar\rho_{t,k}(\theta)\widehat A_t
\right)
\right],
\label{eq:token_ppo}
\end{equation}
where $\epsilon$ is the clipping threshold. Appendix~\ref{sec:ppo_comparison} reports an empirical comparison supporting this backbone choice. These interface and optimization adaptations enable direct RL fine-tuning of the pretrained VLA policy while preserving the original action interface.

\subsection{Failure-Aware Curriculum under Scene Imbalance}

\par The optimization backbone learns from both successful and failed trajectories, while sustained improvement also depends on the training opportunities available across tasks and scenes. Although event-based penalties provide diverse failure feedback, ordinary task sampling may offer limited opportunities for further training on the associated tasks. Moreover, long-tailed scene distributions make such opportunities particularly scarce in rare scenes. To address these limitations, RefineFly sustains learning from unresolved tasks through a dynamic failure memory within a two-stage curriculum that shifts training toward a more balanced scene distribution.

\paragraph{Dynamic Failure Memory.}
\par Compared with the terminal success reward, event-based failure penalties provide feedback on a broader range of navigation outcomes. Specifically, collisions, stalled motion, persistent movement away from the target, premature landing, and timeouts reflect distinct forms of unsuccessful execution. To incorporate this diverse feedback into policy optimization, RefineFly integrates event-specific rewards and penalties with dense distance progress, defining the rollout reward as
\begin{equation}
r_t={}\operatorname{clip}\!\left(
\kappa_p(D_{t-1}-D_t),
r_{\min}^{\mathrm{prog}},r_{\max}^{\mathrm{prog}}
\right)+\sum_{e\in\mathcal E}R_e\mathbb{I}[e_t=e],
\label{eq:event_aware_reward}
\end{equation}
where $D_t$ denotes the Euclidean distance to the target, $\kappa_p$ scales the progress reward, and $r_{\min}^{\mathrm{prog}}$ and $r_{\max}^{\mathrm{prog}}$ are its clipping bounds. The set $\mathcal{E}=\{\texttt{success},\allowbreak\texttt{collision},\allowbreak\texttt{stuck},\allowbreak\texttt{away},\allowbreak\texttt{early-stop},\allowbreak\texttt{timeout}\}$ contains the navigation events, and $R_e$ specifies the reward or penalty associated with event $e$. Standard generalized advantage estimation (GAE)~\citep{Schulman2016GAE} uses these rewards and the value estimates $V_\phi(h_t)$ to compute the action-level advantages $\widehat A_t$ for token-level PPO.

\par To sustain learning from these diverse failure signals across rollouts, RefineFly maintains a dynamic failure memory $\mathcal M_u=\{\xi_i\}$ that evolves with online training. Specifically, each task record is represented as
\begin{equation}
\xi_i=(\omega_i,c_i,f_i,\sigma_i,n_i),
\label{eq:failure_memory}
\end{equation}
where the entries denote the task initialization, scene, failure type, status, and unsuccessful retry count, respectively.
Initially, the memory contains failed tasks collected with the initial policy. When an unresolved record is selected, the current policy generates a complete new trajectory from $\omega_i$. After each attempt, a successful task is marked as solved and excluded from further selection, while $n_i$ is incremented after a failed attempt. To limit repeated unsuccessful trials, the record is marked as dropped and excluded when $n_i$ reaches $N_{\max}$. Meanwhile, failures from fresh sampling enter the memory as unresolved records, including previously solved or dropped tasks that fail again.

\paragraph{Two-Stage Scene Curriculum.}
\par To sustain learning from failures while increasing training opportunities for rare scenes, we organize online post-training into a two-stage scene curriculum. Specifically, Stage I follows the empirical scene distribution, while Stage II continues from the final Stage I policy with balanced scene sampling. The overall scene allocations are defined as
\begin{equation}
P_1(c)=\frac{N_c}{\sum_{c'\in\mathcal C}N_{c'}},
\qquad
P_2(c)=\frac{1}{|\mathcal C|},
\label{eq:stage_sampling}
\end{equation}
where $\mathcal C$ denotes the training scenes, and $N_c$ is the number of tasks in scene $c$. For scenes with insufficient tasks, repeated sampling fills the Stage II quota.

\par Both stages reserve a fixed fraction of rollouts for unresolved tasks in the dynamic failure memory while preserving the prescribed scene distribution. Specifically, $q_{s,u}$ denotes the task sampling distribution at iteration $u$ of stage $s\in\{1,2\}$, with scene distribution $P_s$. Within this allocation, a fraction $\eta$ of rollouts is assigned to unresolved tasks in $\mathcal M_u$, and the remaining rollouts use fresh task sampling. The current policy executes the selected tasks, and navigation outcomes update the memory through the same lifecycle. Under the rebalanced distribution of Stage II, tasks in rare scenes receive more training opportunities, allowing additional failures to enter the memory and support subsequent learning. Appendix~A.1 specifies the scene quotas and sampling procedure.

\subsection{Reference-Constrained Policy Updates}
\par Although the failure-aware curriculum strengthens learning from failures and aims to improve generalization under scene imbalance, changes in the training distribution may induce policy drift and weaken previously acquired navigation capabilities. To limit this drift, we adopt reference-policy Kullback-Leibler (KL) regularization. Specifically, the reference policy $\pi_{\mathrm{ref}}^{(s)}$ remains frozen throughout each training stage $s\in\{1,2\}$. In the two-stage curriculum, Stage I uses the initial VLA policy as $\pi_{\mathrm{ref}}^{(1)}$ and Stage II uses the final Stage I policy as $\pi_{\mathrm{ref}}^{(2)}$ to anchor adaptation to the rebalanced scene distribution.

\par The reference penalty measures policy deviation at valid action-token positions. Specifically, let $p_{\theta,t,k}(\cdot)=\pi_\theta(\cdot\mid h_t,z_{t,<k})$ and $p_{\mathrm{ref},t,k}^{(s)}(\cdot)=\pi_{\mathrm{ref}}^{(s)}(\cdot\mid h_t,z_{t,<k})$ denote the corresponding token distributions. Using the mask and normalization defined in the optimization backbone, the reference loss is given by
\begin{equation}
\mathcal L_{\mathrm{KL}}^{(s)}
=\mathbb E_t\!\left[
\frac{1}{M_t}\sum_{k=1}^{K}m_{t,k}
D_{\mathrm{KL}}\!\left(
p_{\theta,t,k}\,\|\,p_{\mathrm{ref},t,k}^{(s)}
\right)
\right].
\label{eq:reference_kl}
\end{equation}

\par The final objective combines the policy, value, and reference terms using trajectories collected under the failure-aware curriculum. Specifically, with task initializations sampled according to $q_{s,u}$, the loss is given by
\begin{equation}
\mathcal L(\theta,\phi)
=-J_{\mathrm{PPO}}^{\mathrm{token}}(\theta)
+c_v\mathcal L_V(\phi)
+\beta\mathcal L_{\mathrm{KL}}^{(s)},
\label{eq:overall_loss}
\end{equation}
where $c_v$ weights value regression and $\beta$ controls the reference penalty. Thus, the scene curriculum specifies the overall scene distribution, while token-level PPO and reference regularization govern policy updates at valid action-token positions.

\section{Experiments}
\label{sec:experiments}

\subsection{Experimental Setup}
\label{sec:setup}

\paragraph{Dataset.}
\par We evaluate RefineFly on the TravelUAV dataset~\citep{traveluav} under the \textit{UAV-Need-Help} setting adopted by AerialVLA. Specifically, we use 7922 training trajectories and evaluate on 1418 Seen, 958 Unseen Map, and 629 Unseen Object trajectories. Following the benchmark settings, trajectories shorter than 250 meters are classified as easy, and the remainder as hard. We report results on the full test set and both difficulty subsets for each split. The training set exhibits a long-tailed scene distribution, as detailed in Appendix~\ref{sec:dataset_appendix}.

\paragraph{Metrics.}
\par We evaluate navigation accuracy, task completion, and path efficiency using navigation error (NE), success rate (SR), oracle success rate (OSR), and success weighted by path length (SPL). Specifically, NE measures the final Euclidean distance to the destination. SR measures successful termination within the target region, including a correct \texttt{LAND} output or near-zero movement for 10 consecutive steps. OSR records whether the executed trajectory ever enters that region, while SPL accounts for both task completion and path efficiency. Lower NE and higher SR, OSR, and SPL indicate better performance.

\paragraph{Implementation Details.}
\par RefineFly post-trains the AerialVLA initialization through online interaction while preserving its action interface. Specifically, AerialVLA combines the OpenVLA-7B~\citep{openvla} backbone with a LoRA~\citep{lora} adapter, and post-training optimizes the adapter together with an added value head using simulator-derived rewards. Counting both fresh and memory-selected task executions, the total rollout budget is approximately 30\% of the training-set size. Training uses 8~$\times$ NVIDIA A100 (80GB) GPUs and takes about 21 hours. We implement RefineFly using RLinf~\citep{rlinf}, AirSim~\citep{airsim}, the TravelUAV environment, and the AerialVLA action-token pipeline. Appendix~\ref{sec:training_parameters} provides the remaining training configurations.

\paragraph{Baselines.}
\par We compare RefineFly with heuristic controls, task-specific UAV-VLN methods, generalist navigation models, and an end-to-end VLA baseline, which either rely on fixed policies, specialized planning modules, or behavior-cloning-based training. For all baselines, we report results from the corresponding publications on the same evaluation splits.
\begin{itemize}
    \item \textbf{Heuristic Methods.} Random Action samples controls without using visual or language inputs, while Fixed Action executes predefined commands.
    \item \textbf{Task-Specific UAV-VLN Models.} CMA \citep{Anderson2018} is a recurrent cross-modal navigation baseline, while TravelUAV \citep{traveluav} combines multimodal features with hierarchical trajectory decoders and an external target detector. TravelUAV-DA extends TravelUAV by aggregating additional corrective trajectories.
    \item \textbf{Generalist Navigation Models.} NavFoM \citep{Zhang2025} uses a navigation foundation model with a dedicated trajectory-planning head, and LongFly \citep{longfly} introduces explicit spatiotemporal modeling for long-horizon UAV navigation.
    \item \textbf{End-to-End VLA Baseline.} AerialVLA \citep{aerialvla} autoregressively generates action tokens for UAV control, and its checkpoint initializes RefineFly, making their comparison a direct evaluation of RL post-training.
\end{itemize}

\begin{table}[t]
\caption{Comparison on the Test Seen set. RefineFly reports mean and standard deviation over three evaluation seeds. NE is in meters, and the other metrics are percentages. Bold and underline indicate the best and second-best results, respectively. Shading marks RefineFly. Human performance is for reference only.}
\label{tab:test_seen_new}
\centering
\begingroup
\fontsize{9pt}{10.5pt}\selectfont
\setlength{\tabcolsep}{1.5pt}
\renewcommand{\arraystretch}{0.9}
\begin{tabularx}{\textwidth}{@{}l*{12}{>{\raggedleft\arraybackslash}X}@{}}
\toprule
Method
& \multicolumn{4}{c}{Full}
& \multicolumn{4}{c}{Easy}
& \multicolumn{4}{c}{Hard} \\
\cmidrule(lr){2-5} \cmidrule(lr){6-9} \cmidrule(lr){10-13}
& NE$\downarrow$ & SR$\uparrow$ & OSR$\uparrow$ & SPL$\uparrow$
& NE$\downarrow$ & SR$\uparrow$ & OSR$\uparrow$ & SPL$\uparrow$
& NE$\downarrow$ & SR$\uparrow$ & OSR$\uparrow$ & SPL$\uparrow$ \\
\midrule
Human
& 14.15 & 94.51 & 94.51 & 77.84
& 11.68 & 95.44 & 95.44 & 76.19
& 17.16 & 93.37 & 93.37 & 79.85 \\
\midrule
Random Action
& 222.20 & 0.14 & 0.21 & 0.07
& 142.07 & 0.26 & 0.39 & 0.13
& 320.12 & 0.00 & 0.00 & 0.00 \\
Fixed Action
& 188.61 & 2.27 & 8.16 & 1.40
& 121.36 & 3.48 & 11.48 & 2.14
& 270.69 & 0.79 & 4.09 & 0.49 \\
CMA
& 135.73 & 8.37 & 18.72 & 7.90
& 84.89 & 11.48 & 24.52 & 10.68
& 197.77 & 4.57 & 11.65 & 4.51 \\
TravelUAV-DA
& 98.66 & 17.45 & 48.87 & 15.76
& 66.40 & 20.26 & 51.23 & 18.10
& 138.04 & 14.02 & 45.98 & 12.90 \\
NavFoM
& 93.05 & 29.17 & 49.24 & 25.03
& 58.98 & 32.91 & 53.16 & 27.87
& 143.83 & 23.58 & 43.40 & 20.80 \\
LongFly
& \second{60.02} & 36.39 & \best{65.87} & 31.07
& \second{38.10} & 38.52 & \best{71.90} & 31.24
& \second{85.20} & 33.94 & \second{58.94} & 30.88 \\
AerialVLA
& 65.88 & \second{47.96} & 57.69 & \second{38.54}
& 43.76 & \second{49.30} & 61.30 & \second{37.14}
& 93.16 & \second{46.30} & 53.23 & \second{40.26} \\
\midrule
\rowcolor{RefineFlyRow}
\textbf{RefineFly}
& \meanres{\best{54.96}}
& \meanres{\best{56.33}}
& \meanres{\second{65.40}}
& \meanres{\best{45.98}}
& \meanres{\best{37.60}}
& \meanres{\best{56.28}}
& \meanres{\second{67.39}}
& \meanres{\best{42.91}}
& \meanres{\best{76.36}}
& \meanres{\best{56.38}}
& \meanres{\best{62.94}}
& \meanres{\best{49.75}} \\

\rowcolor{RefineFlyRow}
\textbf{(Ours)} & \stdres{1.19}
& \stdres{0.15}
& \stdres{0.80}
& \stdres{1.13}
& \stdres{1.31}
& \stdres{0.48}
& \stdres{0.98}
& \stdres{0.58}
& \stdres{1.07}
& \stdres{0.88}
& \stdres{0.59}
& \stdres{1.80} \\
\bottomrule
\end{tabularx}
\endgroup

\end{table}

\begin{table}[t]
\caption{Comparison on the Test Unseen Map set. RefineFly reports mean and standard deviation over three evaluation seeds. NE is in meters, and the other metrics are percentages. Bold and underline indicate the best and second-best results, respectively. Shading marks RefineFly.}
\label{tab:test_unseen_map_new}
\centering
\begingroup
\fontsize{9pt}{10.5pt}\selectfont
\setlength{\tabcolsep}{1.5pt}
\renewcommand{\arraystretch}{0.9}
\begin{tabularx}{\textwidth}{@{}l*{12}{>{\raggedleft\arraybackslash}X}@{}}
\toprule
Method
& \multicolumn{4}{c}{Full}
& \multicolumn{4}{c}{Easy}
& \multicolumn{4}{c}{Hard} \\
\cmidrule(lr){2-5} \cmidrule(lr){6-9} \cmidrule(lr){10-13}
& NE$\downarrow$ & SR$\uparrow$ & OSR$\uparrow$ & SPL$\uparrow$
& NE$\downarrow$ & SR$\uparrow$ & OSR$\uparrow$ & SPL$\uparrow$
& NE$\downarrow$ & SR$\uparrow$ & OSR$\uparrow$ & SPL$\uparrow$ \\
\midrule
Random Action
& 202.98 & 0.00 & 0.00 & 0.00
& 158.46 & 0.00 & 0.00 & 0.00
& 265.88 & 0.00 & 0.00 & 0.00 \\
Fixed Action
& 180.47 & 0.52 & 2.61 & 0.39
& 132.89 & 0.89 & 4.28 & 0.67
& 247.72 & 0.00 & 0.25 & 0.00 \\
CMA
& 141.68 & 2.30 & 10.02 & 2.16
& 102.29 & 3.57 & 14.26 & 3.33
& 197.35 & 0.50 & 4.03 & 0.50 \\
TravelUAV
& 138.80 & 4.18 & 20.77 & 3.84
& 102.94 & 4.63 & 22.82 & 4.24
& 189.46 & 3.53 & 17.88 & 3.28 \\
NavFoM
& 125.10 & 6.30 & 18.95 & 5.68
& 102.41 & 6.77 & 20.07 & 6.04
& 170.58 & 5.36 & 15.71 & 4.97 \\
LongFly
& 108.32 & 11.27 & 30.27 & 9.32
& 78.56 & 12.96 & 34.31 & 10.32
& 148.10 & 9.02 & 24.88 & 7.98 \\
AerialVLA
& \second{67.42} & \second{37.58} & \second{52.92} & \second{28.22}
& \second{44.99} & \second{41.89} & \second{58.47} & \second{29.72}
& \second{99.11} & \second{31.49} & \second{45.09} & \second{26.11} \\
\midrule

\rowcolor{RefineFlyRow}
\textbf{RefineFly}
& \meanres{\best{58.88}}
& \meanres{\best{42.97}}
& \meanres{\best{60.09}}
& \meanres{\best{31.83}}
& \meanres{\best{44.02}}
& \meanres{\best{46.88}}
& \meanres{\best{63.04}}
& \meanres{\best{32.44}}
& \meanres{\best{79.88}}
& \meanres{\best{37.45}}
& \meanres{\best{55.92}}
& \meanres{\best{30.98}} \\

\rowcolor{RefineFlyRow}
\textbf{(Ours)} & \stdres{2.09}
& \stdres{1.76}
& \stdres{1.01}
& \stdres{1.25}
& \stdres{0.99}
& \stdres{1.24}
& \stdres{0.37}
& \stdres{0.72}
& \stdres{4.95}
& \stdres{2.65}
& \stdres{2.27}
& \stdres{2.05} \\

\bottomrule
\end{tabularx}
\endgroup

\end{table}

\begin{table}[t]
\caption{Comparison on the Test Unseen Object set. RefineFly reports mean and standard deviation over three evaluation seeds. NE is in meters, and the other metrics are percentages. Bold and underline indicate the best and second-best results, respectively. Shading marks RefineFly.}
\label{tab:test_unseen_object_new}
\centering
\begingroup
\fontsize{9pt}{10.5pt}\selectfont
\setlength{\tabcolsep}{1.5pt}
\renewcommand{\arraystretch}{0.9}
\begin{tabularx}{\textwidth}{@{}l*{12}{>{\raggedleft\arraybackslash}X}@{}}
\toprule
Method
& \multicolumn{4}{c}{Full}
& \multicolumn{4}{c}{Easy}
& \multicolumn{4}{c}{Hard} \\
\cmidrule(lr){2-5} \cmidrule(lr){6-9} \cmidrule(lr){10-13}
& NE$\downarrow$ & SR$\uparrow$ & OSR$\uparrow$ & SPL$\uparrow$
& NE$\downarrow$ & SR$\uparrow$ & OSR$\uparrow$ & SPL$\uparrow$
& NE$\downarrow$ & SR$\uparrow$ & OSR$\uparrow$ & SPL$\uparrow$ \\
\midrule
Random Action
& 260.14 & 0.16 & 0.16 & 0.16
& 174.10 & 0.48 & 0.48 & 0.48
& 302.96 & 0.00 & 0.00 & 0.00 \\
Fixed Action
& 212.84 & 3.66 & 9.54 & 2.16
& 151.66 & 6.70 & 13.88 & 3.72
& 243.29 & 2.14 & 7.38 & 1.38 \\
CMA
& 155.79 & 9.06 & 16.06 & 8.68
& 102.92 & 14.83 & 22.49 & 13.90
& 182.09 & 6.19 & 12.86 & 6.08 \\
TravelUAV
& 118.11 & 22.42 & 46.90 & 20.51
& 86.12 & 24.40 & 49.28 & 22.03
& 134.03 & 21.43 & 45.71 & 19.75 \\
NavFoM
& 108.04 & 29.83 & 47.99 & 27.20
& 70.51 & 32.54 & 50.72 & 29.54
& 133.01 & 28.03 & 46.18 & 25.64 \\
LongFly
& 66.74 & 43.87 & 64.56 & 38.39
& 54.84 & 38.01 & 56.84 & 31.36
& \best{57.07} & 50.25 & \best{74.16} & 45.27 \\
AerialVLA
& \second{61.45} & \second{56.60} & \second{64.86} & \second{46.61}
& \second{45.72} & \second{56.94} & \second{64.11} & \second{43.76}
& 69.27 & \second{56.43} & 65.24 & \second{48.03} \\
\midrule

\rowcolor{RefineFlyRow}
\textbf{RefineFly}
& \meanres{\best{53.42}}
& \meanres{\best{59.72}}
& \meanres{\best{68.31}}
& \meanres{\best{51.34}}
& \meanres{\best{35.49}}
& \meanres{\best{63.80}}
& \meanres{\best{72.09}}
& \meanres{\best{50.97}}
& \meanres{\second{62.34}}
& \meanres{\best{57.70}}
& \meanres{\second{66.43}}
& \meanres{\best{51.52}} \\

\rowcolor{RefineFlyRow}
\textbf{(Ours)} & \stdres{0.97}
& \stdres{0.81}
& \stdres{1.66}
& \stdres{0.77}
& \stdres{1.94}
& \stdres{1.20}
& \stdres{2.41}
& \stdres{1.58}
& \stdres{0.52}
& \stdres{1.45}
& \stdres{1.95}
& \stdres{1.02} \\

\bottomrule
\end{tabularx}
\endgroup

\end{table}

\subsection{Performance Comparison}
\label{sec:performance}

\paragraph{Performance on Seen Environments.} 
\par As shown in Table~\ref{tab:test_seen_new}, RefineFly improves closed-loop navigation over AerialVLA on the test seen set, with larger gains on Hard trajectories across all four metrics. Specifically, Full-set SR increases by \(8.37\) percentage points, and the gain reaches \(10.08\) points on the Hard subset. Moreover, Hard-set SPL improves by \(9.49\) points. The stronger gains on long trajectories are consistent with dynamic failure memory providing repeated opportunities to refine behavior on failed tasks, where local execution errors can prevent eventual completion.

\paragraph{Generalization to Unseen Maps.}
\par As shown in Table~\ref{tab:test_unseen_map_new}, RefineFly improves navigation over AerialVLA on unseen maps, increasing Full-set SR by \(5.39\) percentage points. Unlike the Easy subset, where NE changes only slightly, Hard-set NE decreases by \(19.23\)~m, suggesting improved navigation over long routes, where execution errors can accumulate. Moreover, the concurrent gains in OSR and SR show that RefineFly improves both target-region reachability and successful completion on unseen maps. These gains are consistent with scene rebalancing giving less frequent scenes more training opportunities, while reference regularization limits policy deviation during continued failure learning.

\paragraph{Generalization to Unseen Objects.}
\par As shown in Table~\ref{tab:test_unseen_object_new}, RefineFly achieves the best Full-set results across all four metrics. Specifically, SR reaches \(59.72\%\), and SPL improves by \(4.73\) percentage points over AerialVLA. On Hard trajectories, LongFly achieves lower NE and higher OSR, whereas RefineFly obtains higher SR and SPL. The higher completion metrics are consistent with sustained learning from execution feedback, where penalties for collisions and premature landing may help refine navigation and termination behavior for unfamiliar objects.

\begin{table}[t]
    \caption{Incremental ablation of RefineFly using evaluation seed 1. ID~1 uses token-level PPO, and ID~4 is the full framework. Checkmarks denote enabled components. Values are full-split SR (\%) and its unweighted mean across splits, with absolute changes from AerialVLA in parentheses. Bold and underline mark the best and second-best results. Shading marks RefineFly.}
    \label{tab:component_ablation}
\centering
    \small
    \setlength{\tabcolsep}{1.5pt}
    \renewcommand{\arraystretch}{0.9}

    \begin{tabularx}{\textwidth}{@{}cccc*{4}{>{\centering\arraybackslash}X}@{}}
        \toprule
        & \multicolumn{3}{c}{Framework Components}
        & \multicolumn{4}{c}{Success Rate (\%)} \\
        \cmidrule(lr){2-4}
        \cmidrule(lr){5-8}

        ID
        & \shortstack{Failure\\Memory}
        & \shortstack{Reference\\KL}
        & \shortstack{Two-Stage\\Curriculum}
        & Seen
        & \shortstack{Unseen\\Map}
        & \shortstack{Unseen\\Object}
        & Avg. \\
        \midrule

        $-$
        & $-$
        & $-$
        & $-$
        & 47.96
        & 37.58
        & 56.60
        & 47.38 \\

        \midrule

        1
        & $\times$
        & $\times$
        & $\times$
        & 48.17\,$(+0.21\,\uparrow)$
        & \underline{42.90}\,$(+5.32\,\uparrow)$
        & 56.44\,$(-0.16\,\downarrow)$
        & 49.17\,$(+1.79\,\uparrow)$ \\

        2
        & $\checkmark$
        & $\times$
        & $\times$
        & \underline{56.21}\,$(+8.25\,\uparrow)$
        & 31.21\,$(-6.37\,\downarrow)$
        & 59.78\,$(+3.18\,\uparrow)$
        & 49.07\,$(+1.69\,\uparrow)$ \\

        3
        & $\checkmark$
        & $\checkmark$
        & $\times$
        & 55.01\,$(+7.05\,\uparrow)$
        & 37.47\,$(-0.11\,\downarrow)$
        & \textbf{62.32}\,$(+5.72\,\uparrow)$
        & \underline{51.60}\,$(+4.22\,\uparrow)$ \\

        \rowcolor{RefineFlyRow}
        4
        & $\checkmark$
        & $\checkmark$
        & $\checkmark$
        & \textbf{56.28}\,$(+8.32\,\uparrow)$
        & \textbf{44.89}\,$(+7.31\,\uparrow)$
        & \underline{60.41}\,$(+3.81\,\uparrow)$
        & \textbf{53.86}\,$(+6.48\,\uparrow)$ \\

        \bottomrule
    \end{tabularx}

\end{table}

\begin{table}[t]
    \caption{Scene sampling comparison using full-split SR (\%) with evaluation seed 1. Uniform Sampling uses equal scene quotas in one stage, while Original Distribution retains empirical proportions across two stages. RefineFly shifts from empirical to balanced sampling. Bold and underline mark the best and second-best results. Shading marks RefineFly.}
    \label{tab:sampling_strategy_ablation}
\centering
    \small
    \setlength{\tabcolsep}{3.0pt}
    \renewcommand{\arraystretch}{0.9}

    \begin{tabularx}{\columnwidth}{@{}l*{4}{>{\centering\arraybackslash}X}@{}}
        \toprule
        Sampling Strategy
        & Seen
        & \shortstack{Unseen\\Map}
        & \shortstack{Unseen\\Object}
        & Avg. \\
        \midrule

        Uniform Sampling
        & 50.71
        & \underline{41.23}
        & 52.31
        & 48.08 \\

        Original Distribution
        & \underline{56.13}
        & 40.92
        & \underline{59.14}
        & \underline{52.06} \\

        \rowcolor{RefineFlyRow}
        Two-Stage Curriculum
        & \textbf{56.28}
        & \textbf{44.89}
        & \textbf{60.41}
        & \textbf{53.86} \\

        \bottomrule
    \end{tabularx}

\end{table}

\subsection{Ablation Study}
\label{sec:ablation}

\par We conduct ablation studies to investigate how sustained failure learning interacts with reference regularization and scene sampling across all test splits. Specifically, all ablations use evaluation seed 1. Table~\ref{tab:component_ablation} incrementally adds components to the token-level PPO backbone, while Table~\ref{tab:sampling_strategy_ablation} compares scene sampling schedules. Appendices~\ref{sec:replay_details} and~\ref{sec:ppo_comparison} analyze failure-memory outcomes and compare token-level and sequence-level PPO.

\paragraph{Effect of RL Post-Training and Dynamic Failure Memory.}
\par As shown in Table~\ref{tab:component_ablation}, the average performance improvement gained from using only token-level PPO is relatively limited, and the gains are uneven across different evaluation splits. Compared with AerialVLA, the token-level PPO baseline (ID 1) leaves Seen SR nearly unchanged, improves Unseen Map SR by 5.32 percentage points, and reduces Unseen Object SR by 0.16 points. The average gain of 1.79 percentage points supports policy refinement through online feedback, while the uneven changes motivate examining how failure-learning opportunities are distributed during post-training.

\par Furthermore, dynamic failure memory improves performance on some evaluation splits while exposing a generalization trade-off. Specifically, ID~2 increases Seen and Unseen Object SR by \(8.04\) and \(3.34\) points relative to ID~1, whereas Unseen Map SR decreases by \(11.69\) points. Repeated current-policy rollouts provide fresh feedback on unresolved tasks, supporting further adaptation to previously unsuccessful navigation conditions. However, the unchanged empirical scene allocation does not increase the relative training share of rare scenes. These results suggest that continued task-specific practice can improve navigation in familiar scenes while leaving transfer to unseen maps limited, motivating further reference constraints and scene rebalancing.

\paragraph{Effect of Reference-Policy KL Regularization.}
\par Reference-policy KL regularization partially offsets the generalization loss associated with failure-focused training. As can be seen in Table~\ref{tab:component_ablation}, ID~3 improves Unseen Map SR by \(6.26\) percentage points and average SR by \(2.53\) points relative to ID~2, while Seen SR decreases by \(1.20\) points. Anchoring updates to the stage-initial policy discourages large policy deviations during repeated training on unresolved tasks. The improvement on unseen maps and reduction on seen environments are consistent with tempering task-specific adaptation to support transfer. However, Unseen Map SR remains below the token-level PPO baseline, leaving an unresolved trade-off for the scene curriculum to address.

\paragraph{Effect of the Two-Stage Curriculum.}
\par The two-stage curriculum further improves average performance by changing the scene allocation under which failure learning continues. As shown in Table~\ref{tab:component_ablation}, ID~4 improves Unseen Map and average SR by \(7.42\) and \(2.26\) percentage points relative to ID~3, with a \(1.91\)-point decrease in Unseen Object SR. Balanced sampling in Stage II gives rare scenes more opportunities for both fresh interaction and continued failure learning, offering a plausible explanation for the gain on unseen maps. Furthermore, as illustrated in Table~\ref{tab:sampling_strategy_ablation}, RefineFly outperforms uniform sampling and the two-stage original-distribution baseline on all three splits. Relative to the latter, Unseen Map and average SR increase by \(3.97\) and \(1.80\) points, respectively. These results support a strategy of initial optimization based on empirical distributions followed by scene rebalancing, suggesting that sustained failure learning benefits from how training opportunities are allocated across stages.

\section{Conclusions}
\label{sec:conclusion}

\par We present RefineFly, a failure-aware post-training framework that sustains learning from closed-loop failures through dynamic failure memory within a two-stage scene curriculum built on token-level PPO. With a total rollout budget of about 30\% of the training-set size, RefineFly outperforms all comparison methods across all three splits, improving the SR of AerialVLA by 3.12 to 8.37 percentage points. Beyond these gains, our findings indicate that failure experience provides valuable opportunities for policy improvement, while its benefits vary across evaluation splits under an imbalanced training distribution. Moreover, shifting from empirical to balanced scene sampling improves generalization and outperforms uniform sampling from the outset, highlighting the importance of how training opportunities are distributed when learning from failures. Extending this evaluation to additional benchmarks and real-world flights is an important direction for future work.

\label{sec:main_end}

\bibliography{references}

@InProceedings{Anderson2018,
  author    = {Peter Anderson and Qi Wu and Damien Teney and Jake Bruce and Mark Johnson and Niko Sünderhauf and Ian Reid and Stephen Gould and Anton van den Hengel},
  booktitle = {{P}roc. CVPR},
  title     = {Vision-and-Language Navigation: Interpreting Visually-Grounded Navigation Instructions in Real Environments},
  year      = {2018},
  pages     = {3674--3683},
}

@InProceedings{Krantz2020,
  author    = {Jacob Krantz and Erik Wijmans and Arjun Majumdar and Dhruv Batra and Stefan Lee},
  booktitle = {{P}roc. ECCV},
  title     = {Beyond the Nav-Graph: Vision-and-Language Navigation in Continuous Environments},
  year      = {2020},
  editor    = {Andrea Vedaldi and Horst Bischof and Thomas Brox and Jan{-}Michael Frahm},
  pages     = {104--120},
}

@InProceedings{aerialvln,
  author    = {Shubo Liu and Hongsheng Zhang and Yuankai Qi and Peng Wang and Yanning Zhang and Qi Wu},
  booktitle = {{P}roc. ICCV},
  title     = {AerialVLN: {V}ision-and-Language Navigation for {UAVs}},
  year      = {2023},
  pages     = {15338--15348},
}

@InProceedings{Fan2023,
  author    = {Yue Fan and Winson Chen and Tongzhou Jiang and Chun Zhou and Yi Zhang and Xin Wang},
  booktitle = {{P}roc. ACL},
  title     = {Aerial Vision-and-Dialog Navigation},
  year      = {2023},
  pages     = {3043--3061},
}

@InProceedings{traveluav,
  author    = {Wang, Xiangyu and Yang, Donglin and Kwan, Hohin and Chen, Jinyu and Li, Hongsheng and Liao, Yue and Liu, Si and others},
  booktitle = {{P}roc. ICLR},
  title     = {Towards realistic {UAV} vision-language navigation: {P}latform, benchmark, and methodology},
  year      = {2025},
  pages     = {7292--7310},
}

@Misc{lin2025openvln,
  author        = {Lin, Peican and Sun, Gan and Liu, Chenxi and Li, Fazeng and Ren, Weihong and Cong, Yang},
  title         = {OpenVLN: {O}pen-world Aerial Vision-Language Navigation},
  year          = {2025},
  archiveprefix = {arXiv},
  eprint        = {2511.06182},
  journal       = {arXiv preprint},
}

@Misc{longfly,
  author        = {Jiang, Wen and Wang, Li and Huang, Kangyao and Fan, Wei and Liu, Jinyuan and Liu, Shaoyu and Duan, Hongwei and Xu, Bin and Ji, Xiangyang},
  title         = {LongFly: {L}ong-Horizon {UAV} Vision-and-Language Navigation with Spatiotemporal Context Integration},
  year          = {2025},
  archiveprefix = {arXiv},
  eprint        = {2512.22010},
  journal       = {arXiv preprint},
}

@Misc{aerialvla,
  author        = {Xu, Peng and Deng, Zhengnan and Deng, Jiayan and Gu, Zonghua and Wan, Shaohua},
  title         = {{AerialVLA}: {A} Vision-Language-Action Model for {UAV} Navigation via Minimalist End-to-End Control},
  year          = {2026},
  archiveprefix = {arXiv},
  eprint        = {2603.14363},
  journal       = {arXiv preprint},
}

@InProceedings{openvla,
  author    = {Kim, Moo Jin and Pertsch, Karl and Karamcheti, Siddharth and Xiao, Ted and Balakrishna, Ashwin and Nair, Suraj and Rafailov, Rafael and Foster, Ethan and Lam, Grace and Sanketi, Pannag and others},
  booktitle = {{P}roc. CoRL},
  title     = {OpenVLA: {A}n open-source vision-language-action model},
  year      = {2025},
  pages     = {2679--2713},
}

@InProceedings{Zitkovich2023,
  author    = {Brianna Zitkovich and Tianhe Yu and Sichun Xu and Peng Xu and Ted Xiao and Fei Xia and Jialin Wu and Paul Wohlhart and Stefan Welker and Ayzaan Wahid and Quan Vuong and Vincent Vanhoucke and Huong T. Tran and Radu Soricut and Anikait Singh and Jaspiar Singh and Pierre Sermanet and Pannag R. Sanketi and Grecia Salazar and Michael S. Ryoo and Krista Reymann and Kanishka Rao and Karl Pertsch and Igor Mordatch and Henryk Michalewski and Yao Lu and Sergey Levine and Lisa Lee and Tsang{-}Wei Edward Lee and Isabel Leal and Yuheng Kuang and Dmitry Kalashnikov and Ryan Julian and Nikhil J. Joshi and Alex Irpan and Brian Ichter and Jasmine Hsu and Alexander Herzog and Karol Hausman and Keerthana Gopalakrishnan and Chuyuan Fu and Pete Florence and Chelsea Finn and Kumar Avinava Dubey and Danny Driess and Tianli Ding and Krzysztof Marcin Choromanski and Xi Chen and Yevgen Chebotar and Justice Carbajal and Noah Brown and Anthony Brohan and Montserrat Gonzalez Arenas and Kehang Han},
  booktitle = {{P}roc. CoRL},
  title     = {{RT-2:} Vision-Language-Action Models Transfer Web Knowledge to Robotic Control},
  year      = {2023},
  editor    = {Jie Tan and Marc Toussaint and Kourosh Darvish},
  pages     = {2165--2183},
}

@Misc{Schulman2017,
  author        = {John Schulman and Filip Wolski and Prafulla Dhariwal and Alec Radford and Oleg Klimov},
  title         = {Proximal Policy Optimization Algorithms},
  year          = {2017},
  archiveprefix = {arXiv},
  eprint        = {1707.06347},
  journal       = {arXiv preprint},
}

@InProceedings{Ross2011,
  author    = {St{\'{e}}phane Ross and Geoffrey J. Gordon and Drew Bagnell},
  booktitle = {{P}roc. AISTATS},
  title     = {A Reduction of Imitation Learning and Structured Prediction to No-Regret Online Learning},
  year      = {2011},
  pages     = {627--635},
}

@InProceedings{airsim,
  author    = {Shah, Shital and Dey, Debadeepta and Lovett, Chris and Kapoor, Ashish},
  booktitle = {Field and service robotics: {R}esults of the 11th international conference},
  title     = {Airsim: {H}igh-fidelity visual and physical simulation for autonomous vehicles},
  year      = {2017},
  pages     = {621--635},
}

@Misc{Zhang2025,
  author        = {Jiazhao Zhang and Anqi Li and Yunpeng Qi and Minghan Li and Jiahang Liu and Shaoan Wang and Haoran Liu and Gengze Zhou and Yuze Wu and Xingxing Li and Yuxin Fan and Wenjun Li and Zhibo Chen and Fei Gao and Qi Wu and Zhizheng Zhang and He Wang},
  title         = {Embodied Navigation Foundation Model},
  year          = {2025},
  archiveprefix = {arXiv},
  eprint        = {2509.12129},
  journal       = {arXiv preprint},
}

@InProceedings{xiao2025uav,
  author    = {Xiao, Jianqiang and Sun, Yuexuan and Shao, Yixin and Gan, Boxi and Liu, Rongqiang and Wu, Yanjin and Guan, Weili and Deng, Xiang},
  booktitle = {{P}roc. ACM MM},
  title     = {{UAV}-{ON}: {A} benchmark for open-world object goal navigation with aerial agents},
  year      = {2025},
  pages     = {13023--13029},
}

@InProceedings{gao2025openfly,
  author    = {Gao, Yunpeng and Li, Chenhui and You, Zhongrui and Liu, Junli and Li, Zhen and Chen, Pengan and Chen, Qizhi and Tang, Zhonghan and Wang, Liansheng and Yang, Penghui and others},
  booktitle = {{P}roc. ICLR},
  title     = {OpenFly: {A} comprehensive platform for aerial vision-language navigation},
  year      = {2026},
}

@InProceedings{zhang2025citynavagent,
  author    = {Zhang, Weichen and Gao, Chen and Yu, Shiquan and Peng, Ruiying and Zhao, Baining and Zhang, Qian and Cui, Jinqiang and Chen, Xinlei and Li, Yong},
  booktitle = {{P}roc. ACL},
  title     = {Citynavagent: {A}erial vision-and-language navigation with hierarchical semantic planning and global memory},
  year      = {2025},
  pages     = {31292--31309},
}

@InProceedings{li2025skyvln,
  author    = {Li, Tianshun and Huai, Tianyi and Li, Zhen and Gao, Yichun and Li, Haoang and Zheng, Xinhu},
  booktitle = {{P}roc. IROS},
  title     = {SkyVLN: {V}ision-and-Language Navigation and {NMPC} Control for {UAVs} in Urban Environments},
  year      = {2025},
  pages     = {17199--17206},
}

@Article{Chen_2025,
  author   = {Chen, Guojun and Yu, Xiaojing and Ling, Neiwen and Zhong, Lin},
  journal  = {IEEE Trans. Mob. Comput.},
  title    = {TypeFly: {L}ow-Latency Drone Planning With Large Language Models},
  year     = {2025},
  pages    = {9068–-9079},
  fjournal = {IEEE Transactions on Mobile Computing},
}

@InProceedings{hu2025see,
  author    = {Hu, Chih Yao and Lin, Yang-Sen and Lee, Yuna and Su, Chih-Hai and Lee, Jie-Ying and Tsai, Shr-Ruei and Lin, Chin-Yang and Chen, Kuan-Wen and Ke, Tsung-Wei and Liu, Yu-Lun},
  booktitle = {{P}roc. CoRL},
  title     = {See, Point, Fly: {A} Learning-Free {VLM} Framework for Universal Unmanned Aerial Navigation},
  year      = {2025},
  pages     = {4697--4708},
}

@InProceedings{wang2026uav,
  author    = {Wang, Xiangyu and Yang, Donglin and Liao, Yue and Zheng, Wenhao and Dai, Bin and Li, Hongsheng and Liu, Si and others},
  booktitle = {{P}roc. NeurIPS},
  title     = {UAV-flow colosseo: {A} real-world benchmark for flying-on-a-word {UAV} imitation learning},
  year      = {2026},
  volume    = {38},
}

@Misc{serpiva2025racevla,
  author        = {Serpiva, Valerii and Lykov, Artem and Myshlyaev, Artyom and Khan, Muhammad Haris and Abdulkarim, Ali Alridha and Sautenkov, Oleg and Tsetserukou, Dzmitry},
  title         = {RaceVLA: {VLA}-based racing drone navigation with human-like behaviour},
  year          = {2025},
  archiveprefix = {arXiv},
  eprint        = {2503.02572},
  journal       = {arXiv preprint},
}

@Misc{lykov2025cognitivedrone,
  author        = {Lykov, Artem and Serpiva, Valerii and Khan, Muhammad Haris and Sautenkov, Oleg and Myshlyaev, Artyom and Tadevosyan, Grik and Yaqoot, Yasheerah and Tsetserukou, Dzmitry},
  title         = {Cognitivedrone: {A} {VLA} model and evaluation benchmark for real-time cognitive task solving and reasoning in {UAVs}},
  year          = {2025},
  archiveprefix = {arXiv},
  eprint        = {2503.01378},
  journal       = {arXiv preprint},
}

@InProceedings{cai2025flightgpt,
  author    = {Cai, Hengxing and Dong, Jinhan and Tan, Jingjun and Deng, Jingcheng and Li, Sihang and Gao, Zhifeng and Wang, Haidong and Su, Zicheng and Sumalee, Agachai and Zhong, Renxin},
  booktitle = {{P}roc. EMNLP},
  title     = {Flightgpt: {T}owards generalizable and interpretable {UAV} vision-and-language navigation with vision-language models},
  year      = {2025},
  pages     = {6670--6687},
}

@Misc{rlinf,
  author        = {Yu, Chao and Wang, Yuanqing and Guo, Zhen and Lin, Hao and Xu, Si and Zang, Hongzhi and Zhang, Quanlu and Wu, Yongji and Zhu, Chunyang and Hu, Junhao and others},
  title         = {RLinf: {F}lexible and Efficient Large-scale Reinforcement Learning via Macro-to-Micro Flow Transformation},
  year          = {2025},
  archiveprefix = {arXiv},
  eprint        = {2509.15965},
  journal       = {arXiv preprint},
}

@InProceedings{lora,
  author    = {Edward J. Hu and Yelong Shen and Phillip Wallis and Zeyuan Allen{-}Zhu and Yuanzhi Li and Shean Wang and Lu Wang and Weizhu Chen},
  booktitle = {{P}roc. ICLR},
  title     = {{LoRA}: {L}ow-Rank Adaptation of Large Language Models},
  year      = {2022},
}

@InProceedings{Schulman2016GAE,
  author    = {John Schulman and Philipp Moritz and Sergey Levine and Michael I. Jordan and Pieter Abbeel},
  booktitle = {{P}roc. {ICLR}},
  title     = {High-Dimensional Continuous Control Using Generalized Advantage Estimation},
  year      = {2016},
}

@InProceedings{Ding2026HETT,
  author    = {Xichen Ding and Jianzhe Gao and Cong Pan and Wenguan Wang and Jie Qin},
  booktitle = {{P}roc. {AAAI}},
  title     = {History-Enhanced Two-Stage Transformer for Aerial Vision-and-Language Navigation},
  year      = {2026},
  pages     = {18225-18233},
  volume    = {40},
}

@InProceedings{Ning2026LookasideVLN,
  author    = {Ning, Yuwei and Zhao, Ganlong and Qin, Yipeng and Liu, Si and Liu, Yang and Lin, Liang and Li, Guanbin},
  booktitle = {{P}roc. CVPR},
  title     = {LookasideVLN: Direction-Aware Aerial Vision-and-Language Navigation},
  year      = {2026},
  pages     = {32441-32450},
}

@InProceedings{Fan2026HTNav,
  author    = {Fan, Chengjie and Pan, Cong and Liu, Zijian and Liu, Ningzhong and Qin, Jie},
  booktitle = {{P}roc. CVPR},
  title     = {{HTNav}: {A} Hybrid Navigation Framework with Tiered Structure for Urban Aerial Vision-and-Language Navigation},
  year      = {2026},
  pages     = {10976-10985},
}

@InProceedings{Blukis2020UAV,
  author    = {Valts Blukis and Yannick Terme and Eyvind Niklasson and Ross A. Knepper and Yoav Artzi},
  booktitle = {{P}roc. CoRL},
  title     = {Learning to Map Natural Language Instructions to Physical Quadcopter Control using Simulated Flight},
  year      = {2019},
  pages     = {1415--1438},
  volume    = {100},
}

@Misc{zhang2025spatialsky,
  author        = {Lingfeng Zhang and Yuchen Zhang and Hongsheng Li and Haoxiang Fu and Yingbo Tang and Hangjun Ye and Long Chen and Xiaojun Liang and Xiaoshuai Hao and Wenbo Ding},
  title         = {Is your {VLM} Sky-Ready? {A} Comprehensive Spatial Intelligence Benchmark for {UAV} Navigation},
  year          = {2025},
  archiveprefix = {arXiv},
  eprint        = {2511.13269},
  journal       = {arXiv preprint},
}

@Misc{ferrag2025uavbench,
  author        = {Mohamed Amine Ferrag and Abderrahmane Lakas and Merouane Debbah},
  title         = {{UAVBench}: {A}n Open Benchmark Dataset for Autonomous and Agentic {AI} {UAV} Systems via {LLM}-Generated Flight Scenarios},
  year          = {2025},
  archiveprefix = {arXiv},
  eprint        = {2511.11252},
  journal       = {arXiv preprint},
}
\bibliographystyle{iclr2027_conference}
\clearpage
\appendix
\section*{Appendix}
\par This supplementary material details the scene distribution, training configuration, dynamic failure memory, and additional analyses of RefineFly. Section~\ref{sec:settings} specifies the data and optimization settings. Section~\ref{sec:replay_details} describes navigation events, memory outcomes, and paired task-level performance changes. Section~\ref{sec:ppo_comparison} compares token-level and sequence-level PPO, while Section~\ref{sec:qualitative} presents a representative navigation case.

\section{Detailed Experimental Settings}
\label{sec:settings}

\par This section reports the training-set composition and optimization configuration used for RefineFly.

\subsection{Dataset}
\label{sec:dataset_appendix}

\paragraph{Benchmarks for Aerial Agents.}
\par Existing benchmarks for aerial agents and UAV-VLN cover complementary levels of perception, reasoning, and control. AerialVLN \citep{aerialvln} and AVDN \citep{Fan2023} established outdoor instruction-following and dialog-conditioned navigation, while UAV-ON \citep{xiao2025uav} extended evaluation to open-world object-goal search. More recent benchmarks emphasize either data scale or high-level cognition, with OpenFly \citep{gao2025openfly} providing diverse trajectories based on predefined flight-action primitives, SpatialSky-Bench \citep{zhang2025spatialsky} assessing aerial spatial intelligence, and UAVBench \citep{ferrag2025uavbench} evaluating mission-level reasoning.
\par Our study requires closed-loop interaction with executable flight actions and explicit navigation feedback. Accordingly, we adopt TravelUAV \citep{traveluav}, which provides target-oriented navigation with continuous trajectories and evaluation splits covering seen scenes, unseen maps, and unseen objects. Moreover, TravelUAV provides the training and evaluation setting for AerialVLA \citep{aerialvla}, whose end-to-end VLA policy maps onboard observations and linguistic prompts to continuous 3-DoF controls and an intrinsic landing decision. This alignment supports evaluating online post-training in the environment and action space used by the behavior-cloned initialization.

\paragraph{Training-Set Distribution.}
\par
\begin{figure}[t]
    \centering
    \includegraphics[width=0.7\linewidth]{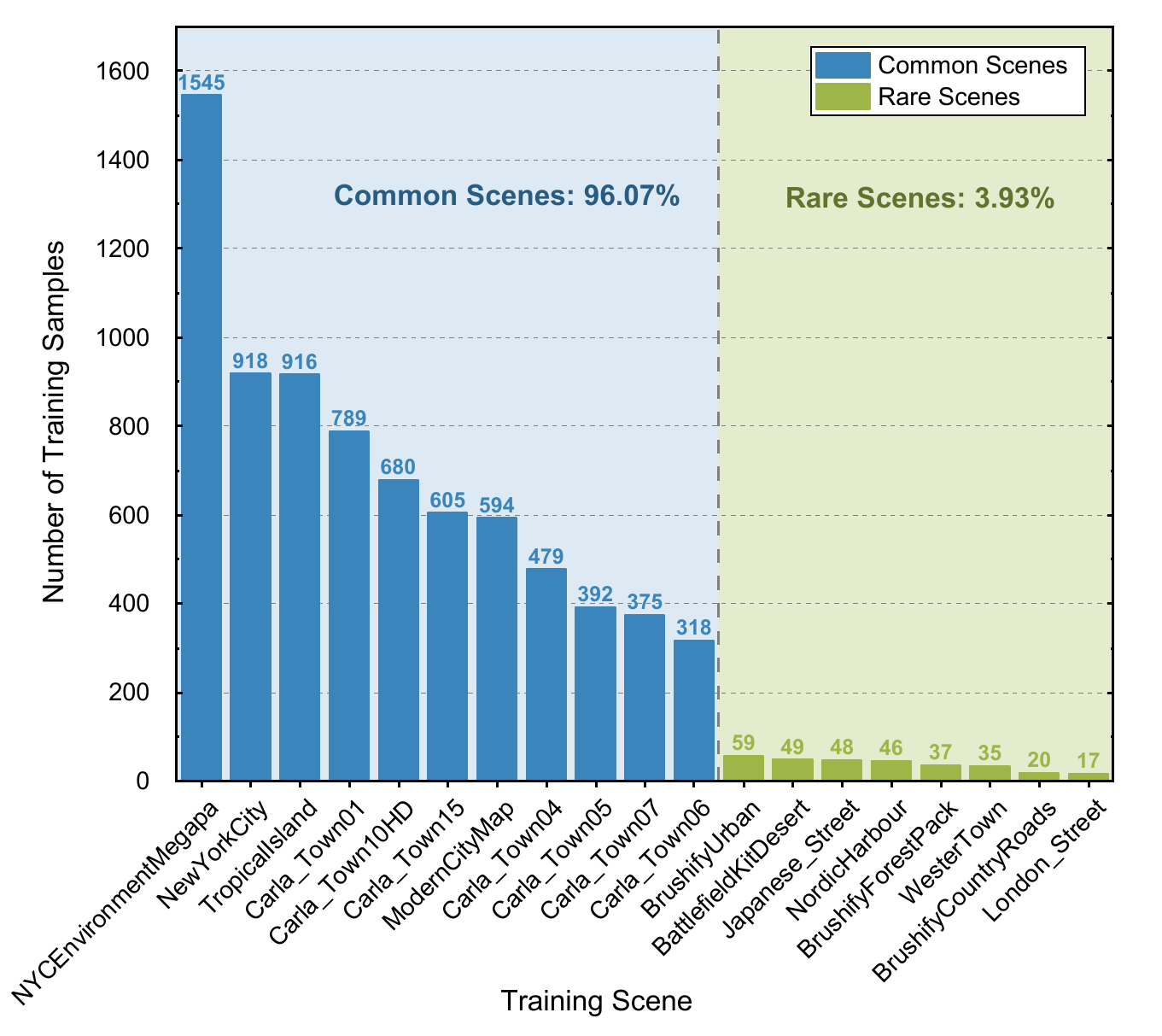}
    \caption{Training sample counts across the 19 TravelUAV scenes used by the \textit{UAV-Need-Help} task.}
    \label{fig:scene_distribution}
\end{figure}

 The TravelUAV training set exhibits a pronounced imbalance across its 19 scenes. Specifically, Fig.~\ref{fig:scene_distribution} reports 7922 training samples, of which 7611 belong to 11 common scenes and account for 96.07\% of the total. The remaining 311 samples span 8 rare scenes and account for only 3.93\%. Furthermore, the largest scene contains 1545 samples, while the rare-scene counts range from 17 to 59. We define the rare-scene set as
\begin{equation}
\mathcal{R}=\{c\in\mathcal{C}\mid N_c<100\},
\label{eq:rare_scene_definition}
\end{equation}
where $N_c$ denotes the number of training samples associated with scene $c$.

\paragraph{Two-Stage Scene Sampling.}
\par The two-stage curriculum changes the overall scene allocation while retaining failure-memory sampling within each stage. Specifically, Stage I follows the empirical scene distribution shown in Fig.~\ref{fig:scene_distribution}, while Stage II assigns an equal quota of 100 selections to each scene. Within these scene allocations, a fraction $\eta$ of rollouts is assigned to unresolved tasks in dynamic failure memory, and the remainder uses fresh task sampling. For fresh sampling, tasks are selected without replacement while unused tasks remain available, followed by sampling with replacement when needed to fill the allocation. Consequently, a complete Stage II quota cycle contains 1900 selections, including 800 from rare scenes. The rare-scene share thus increases from 3.93\% under the empirical distribution to 42.11\%, providing more opportunities for both fresh interaction and continued learning from failures in these scenes.

\begin{table}[t]
\caption{Training and evaluation parameters for RefineFly.}
\label{tab:training_parameters}
\centering
\small
\setlength{\tabcolsep}{4pt}
\renewcommand{\arraystretch}{0.92}
\begin{tabular}{p{0.5\linewidth}p{0.43\linewidth}}
\toprule
Parameter & Value \\
\midrule
\multicolumn{2}{l}{\textbf{Model and Optimization}} \\
\midrule
LoRA rank $r$ & 64 \\
LoRA scaling $\alpha$ & 128 \\
LoRA dropout & 0.05 \\
Policy learning rate & $5\times10^{-5}$ \\
Value head learning rate & $1\times10^{-4}$ \\
Optimizer & AdamW \\
Weight decay & 0.03 \\
Gradient norm clipping & 1.0 \\
Learning rate schedule & Cosine \\
Warm-up steps & 10 \\
Schedule decay factor & 0.01 \\
\midrule
\multicolumn{2}{l}{\textbf{Rollout and Sampling}} \\
\midrule
Stage I rollout epochs & 100 \\
Stage II rollout epochs & 100 \\
Global batch size & 320 \\
Mini-batch size & 8 \\
Sampling temperature & 0.6 \\
Maximum action steps & 200 \\
AirSim \texttt{ClockSpeed} & 10 \\
\midrule
\multicolumn{2}{l}{\textbf{PPO and Dynamic Failure Memory}} \\
\midrule
Discount factor $\gamma$ & 0.995 \\
GAE parameter $\lambda$ & 0.97 \\
PPO clipping threshold $\epsilon$ & 0.2 \\
Value loss coefficient $c_v$ & 0.1 \\
KL coefficient $\beta$ & 0.03 \\
Memory sampling ratio $\eta$ & 0.2 \\
Maximum unsuccessful retries $N_{\max}$ & 2 \\
\midrule
\multicolumn{2}{l}{\textbf{Reward Configuration}} \\
\midrule
Progress reward & $0.02$ per meter, capped at $0.06$ per step \\
Distance increase penalty & $0.02$ per meter, capped at $0.06$ in magnitude \\
Successful landing reward & 3.0 \\
Collision penalty & $-2.0$ \\
Timeout penalty & $-1.0$ \\
Early stop penalty & $-0.8$ \\
Stuck penalty & $-1.2$ \\
Continuous away penalty & $-1.5$ \\
Invalid action penalty & $-0.5$ \\
\bottomrule
\end{tabular}

\end{table}

\subsection{Training and Evaluation Parameters}
\label{sec:training_parameters}

\par RefineFly uses the same optimizer, rollout, and reward configurations in both curriculum stages. Specifically, online post-training continues to optimize the existing adapter with its original LoRA configuration, while the scene distribution and frozen reference policy change between stages as described in the main text. Table~\ref{tab:training_parameters} summarizes the training and evaluation parameters.

\par The online rollout budget counts both fresh task executions and executions selected from dynamic failure memory. Across both stages, fresh task executions amount to approximately 25\% of the training-set size. With memory-selected executions accounting for $\eta=0.2$ of all sampled episodes, the total is $0.25/(1-\eta)=31.25\%$ of the training-set size, reported as approximately 30\%. This ratio measures task executions relative to the number of training tasks, including repeated executions.

\par The main comparison reports RefineFly performance over three evaluation seeds, while ablations use a single evaluation seed. Specifically, the main results give the mean and standard deviation for seeds 1, 42, and 12345. The evaluated post-training configurations in the ablations and corresponding supplementary analyses use seed 1.

\section{Dynamic Failure Memory Details}
\label{sec:replay_details}

\par This section describes the navigation feedback and observed outcomes associated with dynamic failure memory. Specifically, Section~\ref{sec:event_types} defines the events, Section~\ref{sec:failure_distribution} reports their frequencies, and Section~\ref{sec:replay_effect} analyzes the memory lifecycle and task-level changes in evaluation performance.

\subsection{Event Types}
\label{sec:event_types}
\par Navigation events distinguish successful completion from different forms of unsuccessful execution and determine the associated reward feedback.
\paragraph{Success.}
A success event occurs when the UAV is within 20~meters of the target object and either executes \texttt{LAND} or exhibits near-zero movement for 10 consecutive time steps.
\paragraph{Collision.}
A collision event occurs when the simulator reports a physical collision or depth observations indicate that the UAV is too close to an obstacle.
\paragraph{Stuck.}
A stuck event occurs when the UAV exhibits near-zero movement for 10 consecutive steps outside the target region.
\paragraph{Continuous away.}
A continuous-away event occurs when the distance to the target object increases for 10 consecutive steps.
\paragraph{Early stop.}
An early-stop event occurs when the policy issues \texttt{LAND} outside the target region. 
\paragraph{Timeout.}
A timeout event occurs when an episode reaches the 200-step horizon without success.

\par Invalid actions receive a separate format-related penalty. Specifically, an output sequence that cannot be decoded into a valid UAV command produces a safe no-op and receives the penalty in Table~\ref{tab:training_parameters}.

\begin{figure}[t]
    \centering
    \includegraphics[width=0.55\linewidth]{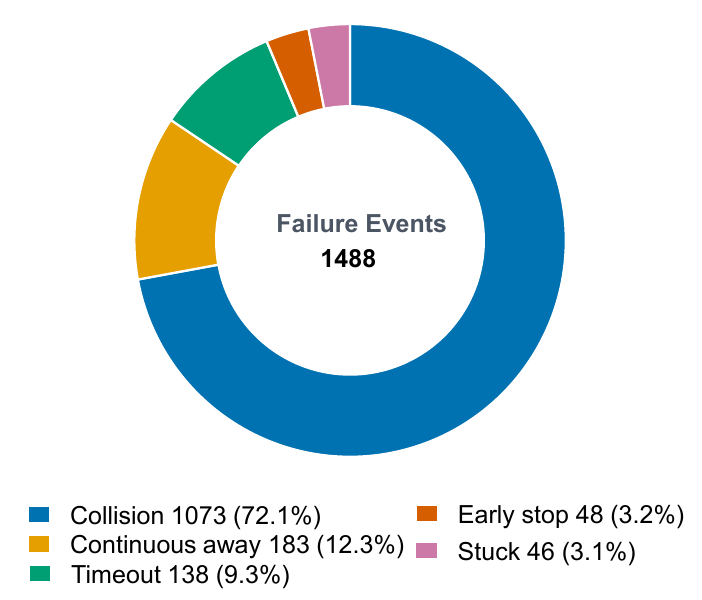}
    \caption{Distribution of 1488 closed-loop navigation failure events recorded during RefineFly training. Collisions account for 1073 events, with 415 events across the other four failure types. Invalid action outputs are excluded.}
    \label{fig:failure_distribution}
\end{figure}

\subsection{Failure Distribution}
\label{sec:failure_distribution}

\par The recorded failures span several navigation difficulties, with collisions accounting for the largest share. Specifically, Fig.~\ref{fig:failure_distribution} summarizes 1488 events from fresh and memory-selected task executions. Collisions account for 1073 events, or 72.1\%, followed by continuous-away and timeout events with 183 and 138 cases, or 12.3\% and 9.3\%. Early-stop and stuck events contribute 48 and 46 cases, together accounting for 6.3\%. These events provide feedback on obstacle avoidance, directional progress, and task termination beyond the terminal success signal. To provide unresolved tasks from the start of post-training, we initialize the memory using failures collected with the initial policy and subsequently update its records using navigation outcomes.

\begin{table}[t]
    \caption{Paired task-level outcomes for token-level PPO (ID~1) and PPO with dynamic failure memory (ID~2) on identical evaluation tasks. $S$ and $F$ denote success and failure. The $F\rightarrow S$ rate is $n_{F\rightarrow S}/(n_{F\rightarrow S}+n_{F\rightarrow F})$, and the $S\rightarrow F$ rate is $n_{S\rightarrow F}/(n_{S\rightarrow S}+n_{S\rightarrow F})$. The net change is $\Delta\mathrm{SR}=100(n_{F\rightarrow S}-n_{S\rightarrow F})/N$. All Splits pools the counts over $N=3005$ tasks and recomputes each rate using its corresponding denominator.}
    \label{tab:paired_replay_transitions}
\centering
    \small
    \setlength{\tabcolsep}{3.0pt}
    \renewcommand{\arraystretch}{1.0}

    \begin{tabular*}{\textwidth}{
        @{\extracolsep{\fill}}
        l cccc cccc
        @{}
    }
        \toprule
        &
        \multicolumn{4}{c}{Paired Outcome Transitions} &
        \multicolumn{4}{c}{Outcome Changes} \\
        \cmidrule(lr){2-5}
        \cmidrule(lr){6-9}

        Evaluation Split
        & $S\!\rightarrow\!S$
        & $S\!\rightarrow\!F$
        & $F\!\rightarrow\!S$
        & $F\!\rightarrow\!F$
        & \shortstack{Net\\Successes}
        & \shortstack{$F\rightarrow S$\\Rate (\%)}
        & \shortstack{$S\rightarrow F$\\Rate (\%)}
        & \shortstack{$\Delta$SR\\(pp)} \\
        \midrule

        Seen
        & 493
        & 190
        & 304
        & 431
        & $+114$
        & 41.36
        & 27.82
        & $+8.04$ \\

        Unseen Map
        & 185
        & 226
        & 114
        & 433
        & $-112$
        & 20.84
        & 54.99
        & $-11.69$ \\

        Unseen Object
        & 268
        & 87
        & 108
        & 166
        & $+21$
        & 39.42
        & 24.51
        & $+3.34$ \\

        \midrule

        All Splits
        & 946
        & 503
        & 526
        & 1,030
        & $+23$
        & 33.80
        & 34.71
        & $+0.77$ \\

        \bottomrule
    \end{tabular*}

\end{table}

\begin{figure}[t]
    \centering
    \includegraphics[width=0.67\linewidth]{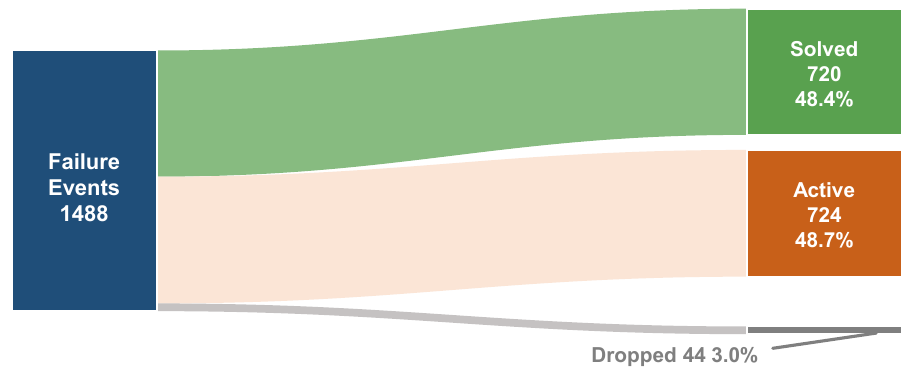}
    \caption{Final states of 1488 failure-memory records over the recorded training interval. The 720 resolved records include successful executions from both memory selection and fresh sampling of active tasks. Another 724 records remain active, while 44 are dropped after two unsuccessful retries.}
    \label{fig:replay_outcomes}
\end{figure}

\begin{figure}[t]
    \centering
    \includegraphics[width=0.67\linewidth]{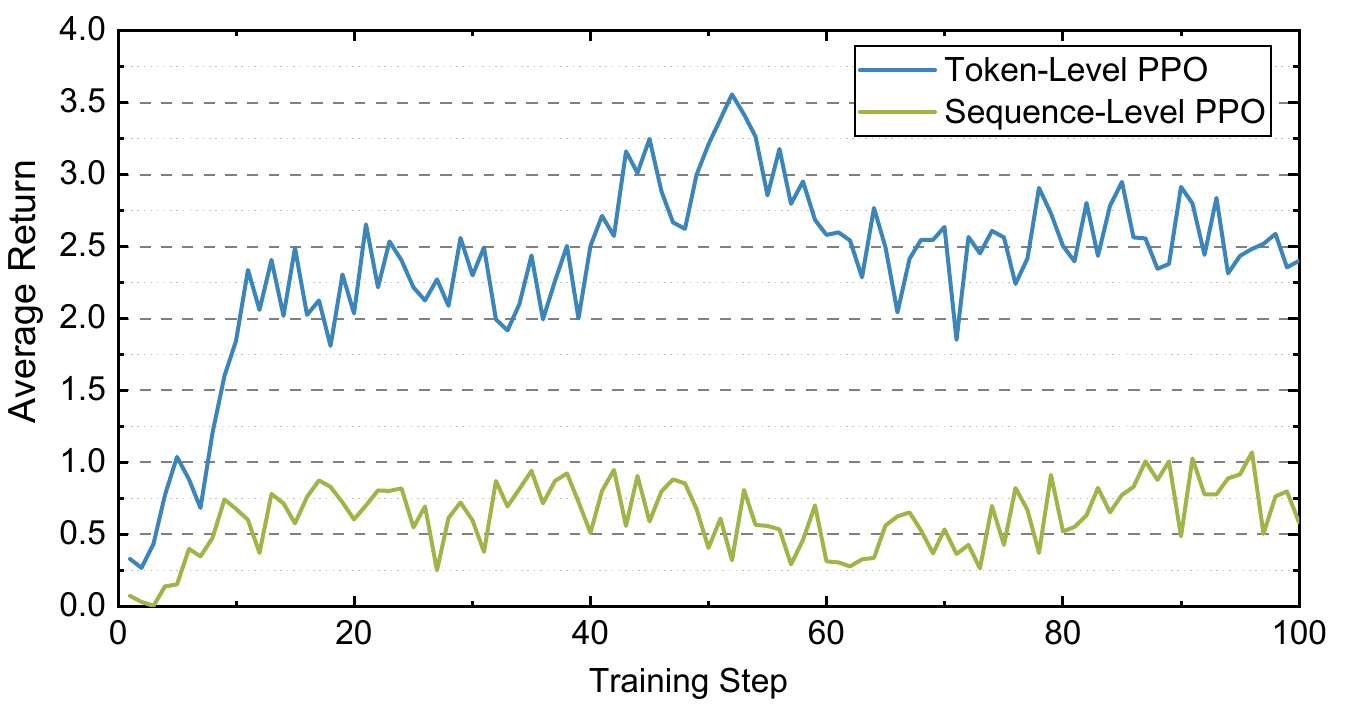}
    \caption{Average return of token-level PPO and the sequence-level baseline over 100 training steps.}
    \label{fig:ppo_convergence}
\end{figure}

\begin{table}[t]
    \caption{Full-split SR (\%) for sequence-level and token-level PPO on the three TravelUAV evaluation splits. Token-level PPO corresponds to ID~1 in the main-paper component ablation. Bold indicates the higher result in each row.}
    \label{tab:ppo_sr_comparison}
\centering
    \small
    \setlength{\tabcolsep}{4pt}
    \renewcommand{\arraystretch}{0.9}
    \begin{tabular*}{\columnwidth}{@{\extracolsep{\fill}}lcc@{}}
        \toprule
        Evaluation Split
        & \shortstack{Sequence-Level\\PPO}
        & \shortstack{Token-Level\\PPO} \\
        \midrule
        Seen          & 41.96 & \textbf{48.17} \\
        Unseen Map    & 37.79 & \textbf{42.90} \\
        Unseen Object & 48.81 & \textbf{56.44} \\
        \bottomrule
    \end{tabular*}

\end{table}

\subsection{Observed Memory Outcomes}
\label{sec:replay_effect}

\par \par Fig.~\ref{fig:replay_outcomes} summarizes the final states of 1488 failure entries. The observed memory states summarize task outcomes from both sampling sources over the recorded interval. Specifically, Fig.~\ref{fig:replay_outcomes} reports 720 resolved, 724 active, and 44 dropped records out of 1488. The resolved count includes active tasks completed successfully through fresh sampling as well as memory-selected executions. Active records remain eligible for further training, while dropped records have exhausted the retry limit. Together, these states characterize the memory lifecycle under the available interaction budget.

\par Dynamic failure memory yields more favorable paired outcomes on Seen and Unseen Object, supporting the value of sustained training on unresolved tasks. Specifically, Table~\ref{tab:paired_replay_transitions} compares the two policies on identical evaluation tasks. On Seen, 304 tasks change from failure under ID~1 to success under ID~2, while 190 change from success to failure, yielding 114 net successes and an 8.04-point SR increase. Similarly, Unseen Object has 108 failure-to-success transitions and 87 success-to-failure transitions, yielding 21 net successes and a 3.34-point SR increase. These gains are consistent with repeated current-policy interaction providing further learning opportunities under previously unsuccessful navigation conditions.

\par The paired outcomes also reveal a trade-off in transfer to unseen maps under the empirical scene distribution. Specifically, Unseen Map has 114 failure-to-success transitions and 226 success-to-failure transitions, yielding 112 net lost successes and an 11.69-point SR decrease. Across all 3005 tasks, the net gain is 23 successes, or 0.77 percentage points. This pooled change weights splits by task count, whereas the main-paper ablation reports an unweighted mean across splits, which decreases by 0.10 points from ID~1 to ID~2. Together, these results indicate that sustained failure learning can improve some navigation outcomes while weakening transfer to new maps. Since memory sampling preserves the empirical scene allocation in this comparison, rare scenes retain a limited share of training opportunities, motivating reference constraints and the scene rebalancing evaluated in the main text.

\begin{figure}[t]
    \centering
    \includegraphics[width=0.99\textwidth]{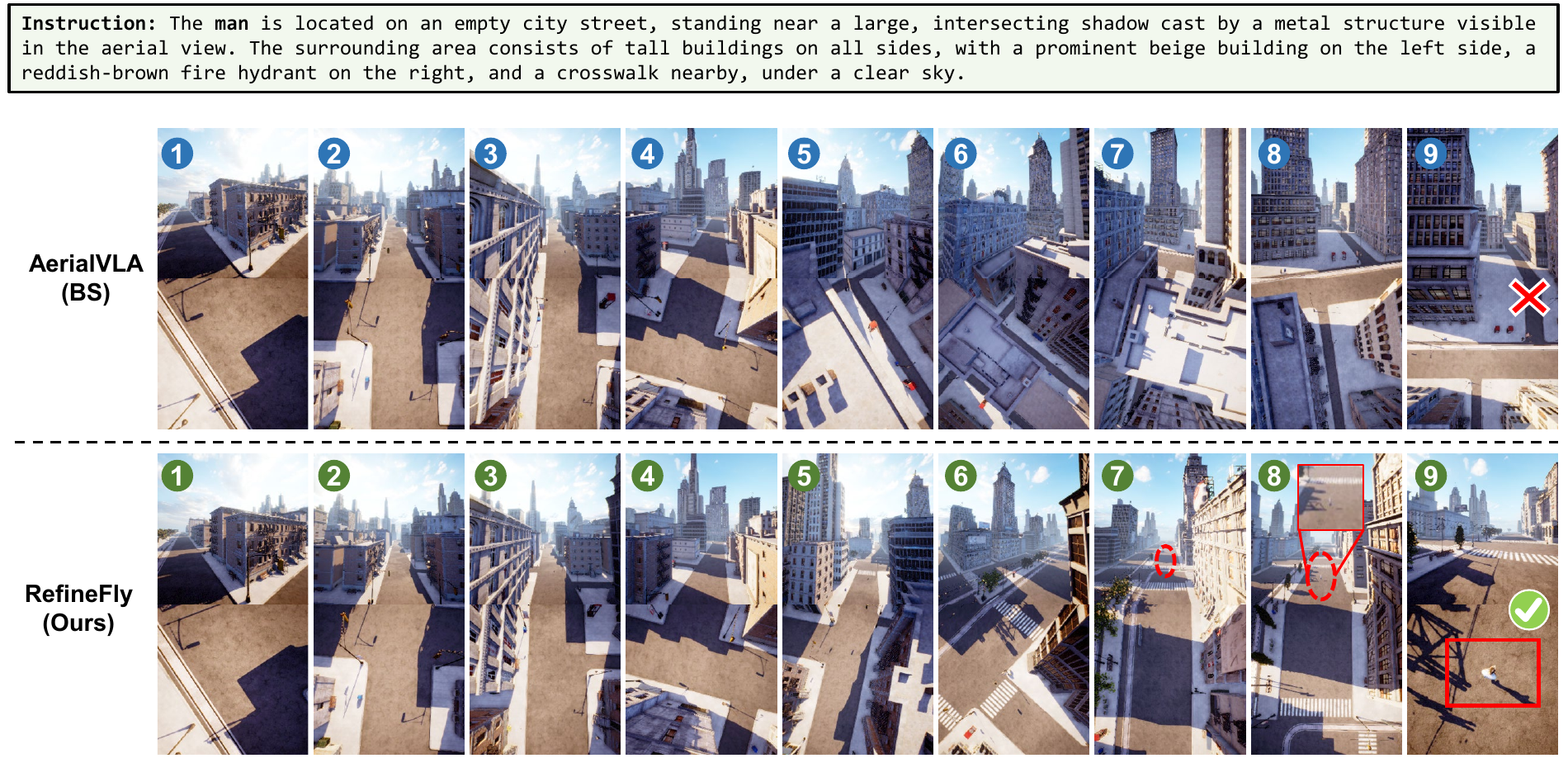}
    \caption{Qualitative navigation comparison under the same instruction and initial observation. The top and bottom rows show key observations from AerialVLA and RefineFly, respectively. Numbers indicate their order within each trajectory.}
    \label{fig:qualitative_comparison}
\end{figure}

\section{Comparison of Token-Level and Sequence-Level PPO}
\label{sec:ppo_comparison}

\par This section provides empirical support for token-level PPO as the optimization backbone of RefineFly. Specifically, the sequence-level alternative computes one importance ratio from the joint likelihood of the complete action-token sequence. Fig.~\ref{fig:ppo_convergence} compares training returns, while Table~\ref{tab:ppo_sr_comparison} reports Full-split success rate (SR) for both alternatives. Token-level PPO corresponds to ID~1 in the main-paper component ablation.

\par Token-level PPO achieves higher training returns than the sequence-level alternative in the recorded run. As shown in Fig.~\ref{fig:ppo_convergence}, its average return rises rapidly during the first 15 training steps and subsequently remains mostly between 2.0 and 3.0, with a peak near 3.5. In comparison, the sequence-level return mostly stays below 1.0 without comparable growth, and the gap persists through the later training steps.

\par The higher training returns are accompanied by better navigation performance across all three evaluation splits. Specifically, token-level PPO exceeds the sequence-level alternative by 5.11 to 7.63 percentage points in SR (Table~\ref{tab:ppo_sr_comparison}). Relative to AerialVLA in Table~\ref{tab:component_ablation}, sequence-level PPO reduces mean SR by 4.53 points, while token-level PPO increases it by 1.79 points. These results support the token-level backbone for the subsequent study of failure memory and scene sampling.

\par The different treatment of action-token probability ratios offers a possible explanation for the observed performance gap. Specifically, each action contains three control tokens and an optional \texttt{LAND} token. Sequence-level PPO couples these probabilities into a joint importance ratio, allowing a change at one position to activate clipping of the shared surrogate objective. In comparison, token-level PPO applies clipping separately at each valid position while sharing the action-level advantage across positions. This formulation reduces coupling in the clipping decisions and may retain learning signals at positions whose individual ratios remain within the clipping range.

\section{Qualitative Analysis}
\label{sec:qualitative}

\par This section illustrates differences in closed-loop navigation between AerialVLA and RefineFly under the same instruction and initial observation. Specifically, Fig.~\ref{fig:qualitative_comparison} presents nine key observations from each trajectory. Both policies initially traverse similar city blocks, while their behavior diverges later in the route.

\par As shown in Fig.~\ref{fig:qualitative_comparison}, AerialVLA initially follows a similar route and subsequently moves away from the target area. In comparison, RefineFly proceeds toward the street marked by the crosswalk and intersecting shadow. The target becomes clearly visible in the eighth observation, after which RefineFly continues its descent and reaches the target location in the final observation. This example illustrates improved target approach and task completion following post-training, consistent with the aggregate navigation gains reported in the main text.

\clearpage
\clearpage
\appendix

\end{document}